\documentclass[10pt,twocolumn,letterpaper]{article}

\usepackage{wacv}              % To produce the CAMERA-READY version
\definecolor{wacvblue}{rgb}{0.21,0.49,0.74}
\usepackage[pagebackref,breaklinks,colorlinks,allcolors=wacvblue]{hyperref}

\def\wacvPaperID{3099} % *** Enter the WACV Paper ID here
\def\confName{WACV}
\def\confYear{2026}
\def\task{expressive composition}
\def\Task{Expressive composition}
\def\netname{StickerNet}

\title{Beyond Realism: Learning the Art of Expressive Composition with StickerNet}

\author{Haoming Lu\\
Picsart AI Research\\
Picsart Inc.\\
{\tt\small jszjlhm@gmail.com}
\and
David Kocharian\\
Picsart AI Research\\
Picsart Inc.\\
{\tt\small david.kocharyan@picsart.com}
\and
Humphrey Shi\\
College of Computing\\
Georgia Institute of Technology\\
{\tt\small shi@gatech.edu}
}

\begin{document}
\maketitle
\begin{abstract}
As a widely used operation in image editing workflows, image composition has traditionally been studied with a focus on achieving visual realism and semantic plausibility. However, in practical editing scenarios of the modern content creation landscape, many compositions are not intended to preserve realism. Instead, users of online platforms motivated by gaining community recognition often aim to create content that is more artistic, playful, or socially engaging. Taking inspiration from this observation, we define the \textit{expressive composition} task, a new formulation of image composition that embraces stylistic diversity and looser placement logic, reflecting how users edit images on real-world creative platforms. To address this underexplored problem, we present \textit{StickerNet}, a two-stage framework that first determines the composition type, then predicts placement parameters such as opacity, mask, location, and scale accordingly. Unlike prior work that constructs datasets by simulating object placements on real images, we directly build our dataset from 1.8 million editing actions collected on \textit{Picsart}, an on-line visual creation and editing platform, each reflecting user-community validated placement decisions. This grounding in authentic editing behavior ensures strong alignment between task definition and training supervision. User studies and quantitative evaluations show that \textit{StickerNet} outperforms common baselines and closely matches human placement behavior, demonstrating the effectiveness of learning from real-world editing patterns despite the inherent ambiguity of the task. This work introduces a new direction in visual understanding that emphasizes expressiveness and user intent over realism.
\end{abstract}
    
\section{Introduction}
\label{sec:intro}

\begin{figure*}[htb]
\centering
\includegraphics[width=0.83\textwidth]{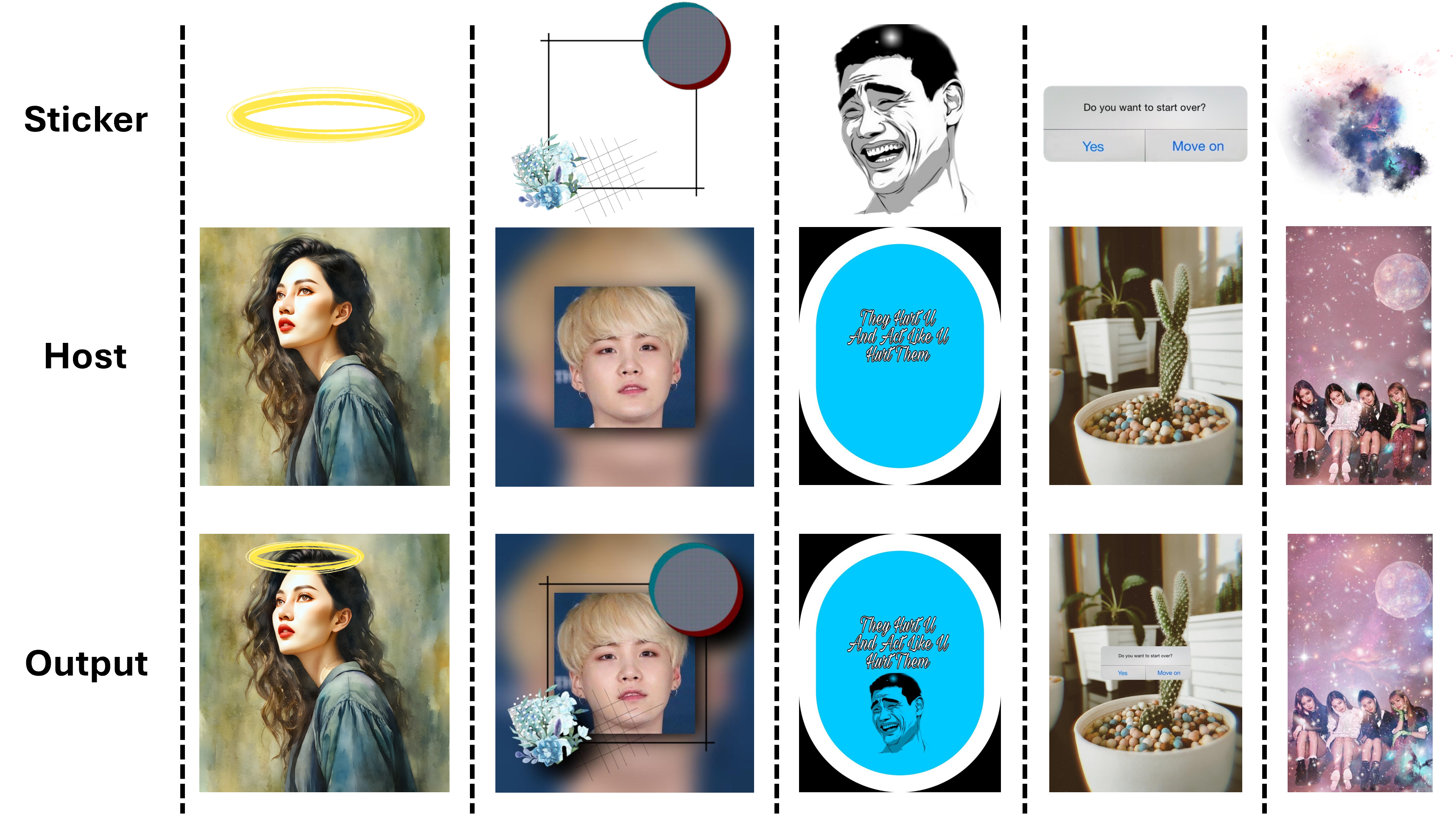}
\vspace{-8pt}
\caption{Examples of \task}
\label{fig:examples}
\vspace{-10pt}
\end{figure*}

Image composition is a fundamental and widely used operation in image editing workflows. Typically, the task involves placing an object onto a background image at a specific scale and position to produce a coherent and visually appealing outcome. In most existing studies~\cite{niu2022fast, lin2018st, zhou2022learning, zhu2023topnet, ye2023efficient, zhang2020learning}, both the background and foreground images are photorealistic, and the main goal is to maintain visual realism and semantic consistency in the final composite. However, in practical editing scenarios, users are often not constrained by realism. Instead, they may aim to create more artistic, playful, or socially engaging content by adding patterned backgrounds, filter effects, or stylistic decorations. As illustrated in \cref{fig:examples}, either the original or the added image (or both) can deviate from a photorealistic style. More fundamentally, the editing intent shifts toward expressiveness and visual impact, rather than realism or semantic alignment. \par

Based on observations of real-world user editing behaviors, we define a new formulation of image composition, called \textbf{\task}. In this setting, the original image is referred to as the \textbf{host image}, and the added element as the \textbf{sticker}. Unlike conventional image composition, which assumes photorealistic content and prioritizes realism or seamless results, \task~exhibits several key differences:

\begin{itemize}
    \item \Task~does not require stylistic consistency or realistic integration between the sticker and the host image, as illustrated in \cref{fig:examples}.
    \item In \task, placement can be guided by the sticker’s own visual or semantic properties, resulting in more aesthetic and visually intentional outcomes, as shown in the last two examples of \cref{fig:examples}.
    \item \Task~incorporates additional interaction modes, including transparency and partial occlusion. As shown in the second and fifth examples of \cref{fig:examples}, stickers can function as backgrounds or overlays.
\end{itemize}

Whereas conventional image composition aims to reconstruct realistic scenes, \task~prioritizes expressiveness and user intent, making the task more ambiguous and difficult to supervise. Traditional image composition models can be trained on synthetic data by segmenting and recombining real images, however, such strategies are not applicable to \task~due to the fundamental differences in task inputs and goals. To support learning under this formulation, we constructed a large-scale dataset of \task~examples derived from popular user edits on a creative image editing platform, capturing real editing intent and community preference. Building on this, we propose \netname, a lightweight and efficient two-stage model designed to reflect the unique characteristics of \task. We evaluated its effectiveness through both quantitative metrics and user studies. The main contributions of this work are as follows:

\begin{itemize}
    \item We define the \textbf{\task} task, a new image composition formulation rooted in real-world user behavior, which prioritizes expressive intent over realism and aligns with practical editing goals in creative contexts.
    \item We construct a large-scale dataset of 1.8 million composition actions from community-validated real editing behaviors, and identify two representative subtypes of composition: filter-style and sticker-style placements.
    \item We present \textbf{\netname}, a lightweight two-stage framework that models composition type and placement jointly, high-quality expressive composition on end-user devices. The model achieves performance comparable to human edits and outperforms standard platform defaults in both quantitative and user study evaluations.
\end{itemize}

\section{Related Works}
\label{sec:related}

\subsection{Image Composition Methods}

Methods for image composition have recently evolved along two primary paradigms. The first, which we refer to as ``dedicated object placement'', focuses on predicting the precise geometric parameters (e.g., bounding boxes) for inserting an object. Another emerging direction treats image composition as a sub-task within text-guided image editing \cite{hui2024hq,sushko2025realedit,zhang2023magicbrush,qin2024think,kawar2023imagic,yang2024editworld,ge2024seed,lu2024can,lu2025trueskin}. A common approach in this paradigm is to fine-tune a pretrained vision-language model (VLM) on a dataset of images paired with text prompts. However, the high computational cost and the prerequisite of explicit text prompts make these models unsuitable for our task. We will discuss this paradigm and its limitations in detail in \cref{sec:related_dataset}. \par

Dedicated object placement methods, which is the focus of our work, can be further categorized by their prediction objectives and strategies. Some studies \cite{song2023objectstitch, ma2024directed} address image composition by enhancing appearance and stylistic consistency based on predefined placement. In contrast, other works focus on predicting object location and transformation using bounding boxes, which aligns more closely with the scope of this work. \par

Based on the prediction strategy, existing object placement approaches can be further divided into two categories. Sparse prediction methods \cite{lee2018context, lin2018st, zhang2020learning, zhou2022sac} estimate object placement by directly predicting bounding box coordinates and scales. In contrast, dense prediction methods \cite{zhu2022gala, zhu2023topnet, niu2022fast, turgutlu2022layoutbert, ye2023efficient} generate a set of possible placements, often via sliding windows, and select the most likely candidate. More specifically, Lee et al. \cite{lee2018context} introduced an end-to-end generative framework to jointly estimate an object's location, scale, and appearance. Generative adversarial networks (GANs) have also been explored for compositional reasoning \cite{zhou2022sac, wang2023gan}, with Zhou et al. \cite{zhou2022learning} proposing a graph-enhanced GAN to further improve spatial and semantic coherence. More recent efforts leverage transformer architectures, including LayoutBERT \cite{turgutlu2022layoutbert}, FTOPNet \cite{ye2023efficient}, and TopNet \cite{zhu2023topnet}. Other directions include U-Net-based frameworks \cite{niu2022fast} and diffusion-based models \cite{liu2024conditional}, which aim to refine spatial arrangement through iterative generation. In addition, PlaceNet \cite{zhang2020learning} adopts a self-learning strategy that combines object detection and inpainting to produce plausible compositions. \par

While prior methods offer valuable insights, they are predominantly designed for photorealistic settings, focusing on maintaining realism and semantic coherence. Many are further limited to specific object types or background domains such as urban scenes~\cite{zhou2022sac}. Although effective in conventional contexts, these assumptions conflict with the objectives of \task, which emphasizes expressiveness and stylistic flexibility. As shown in \cref{sec:exp}, models trained with such constraints show limited performance in \task, highlighting a gap between existing paradigms and the goals of our task.

\subsection{Related Datasets}
\label{sec:related_dataset}

Existing datasets for object composition can be broadly categorized into two groups, as summarized in \cref{tab:dataset}. The first category consists of datasets designed specifically for object placement, such as OPAZ~\cite{qin2024think}, OPA~\cite{liu2021opa}, and ORIDa~\cite{kim2025orida}. These datasets are typically constructed by composing existing foreground objects and background images, with labels indicating whether the result is rational or realistic. However, they suffer from several limitations. First, their scale and diversity are restricted as they are synthesized from real photos, which fails to cover various user scenarios. Second, their objective of achieving semantic realism is often misaligned with the user's goal of creating novel, expressive, and visually engaging content. \par

The second category includes recent instruction-guided image editing datasets, such as MagicBrush~\cite{zhang2023magicbrush}, RealEdit~\cite{sushko2025realedit}, and HQ-Edit~\cite{hui2024hq}. While these datasets often originate from real user samples, they present their own challenges. Their scale is still limited, and the subset of edits involving add-object operations is even smaller. A more significant issue is their reliance on explicit textual instructions, which are not always available in creative scenarios where users edit visually and intuitively. Furthermore, models leveraging this data often require a large multimodal model (LMM) to process language requests, leading to substantial computational costs and latency that restrict their practical application. \par

In contrast, our dataset, which will be detailed in \cref{sec:dataset}, fills a crucial gap in the existing literature. It is large-scale, derived entirely from real user interactions, and critically, does not rely on any additional guidance. By focusing on capturing user intent for expressiveness directly from their actions, our dataset enables the development of efficient, guidance-free models capable of producing creative and compelling compositions.

\begin{table*}[ht]
  \centering
  \begin{tabular}{@{}llll@{}}
    \toprule
    Dataset & Size & Guidance & Target (from original papers)\\
    \midrule
    OPAZ\cite{qin2024think} & 8160 (1390 rational, 6770 irrational) & None & Semantically realism \\
    OPA\cite{liu2021opa} & 73470 (24964 positive, 48506 negative) & None & Rationality \\
    ORIDa\cite{kim2025orida} & 200 objects, 30000 compositions & None & Factual composition \\
    MagicBrush\cite{zhang2023magicbrush} & 10K edits (39\% with add-object) & Text & Instruction-guided editing \\
    RealEdit\cite{sushko2025realedit} & 57K edits & Text & Address user request \\
    HQ-Edit\cite{hui2024hq} & 200K edits (10.7\% with add-object) & Text & Instruction-based editing \\
    \midrule
    Ours & 1881169 real edits & None & Expressiveness\\
    \bottomrule
  \end{tabular}
  \caption{Comparison of our dataset with existing datasets: our work focuses on expressiveness from real user edits without explicit text guidance, filling a key gap.}
  \vspace{-6pt}
  \label{tab:dataset}
\end{table*}

\subsection{Evaluation Metrics}

While several metrics are used to evaluate image composition, many are misaligned with the subjective goals of \task. In this section, we review five common metrics to assess their suitability and justify our choice of a primary evaluation method.

\begin{itemize}
\item \textbf{Accuracy}, as introduced in \cite{zhou2022learning}, uses an auxiliary binary classifier trained on human-labeled data to assess whether a composite image appears reasonable. However, semantic plausibility, as targeted by this metric, is relatively objective. In contrast, \task~aims for more subjective goals, such as aesthetic quality and expressiveness, which cannot be reliably assessed by binary classification. We therefore employ a finer-grained user study to evaluate the outcomes as demonstrated in \cref{sec:exp}.

\item \textbf{Fréchet Inception Distance (FID)} measures the distributional divergence between generated and training images. Although effective in tasks focused on realistic reconstruction, it is less relevant to our task, where the outputs lack a clear distributional target.

\item \textbf{Learned Perceptual Image Patch Similarity (LPIPS)} \cite{zhou2022learning, zhang2018unreasonable} measures output diversity through random sampling. Although diversity is often encouraged in generative tasks, it does not imply better performance in \task, where many cases have only one reasonable placement, as shown in \cref{fig:examples}, columns 1, 2, and 4.

\item \textbf{Intersection over Union (IoU)} measures the overlap between the predicted and ground-truth boxes and is widely used in object detection. In image composition, lower IoU may result from ambiguity rather than incorrect prediction, but higher IoU generally indicates a more accurate placement. Therefore, IoU is adopted as a metric to evaluate placement quality for \task.
 
\item \textbf{User Preference} also serves as a direct form of evaluation, beyond its use in training an auxiliary classifier for an \textbf{Accuracy} metric. Given that the core objective of \task~is to align with human notions of expressiveness and visual appeal, direct human evaluation is the most reliable way to measure performance. We therefore adopt user preference as our primary metric as detailed in \cref{sec:exp}.

\end{itemize}

\section{A Large-Scale Dataset of Real-World Expressive Compositions}

\label{sec:dataset}

\begin{figure*}[htb]
\centering
\includegraphics[width=\textwidth]{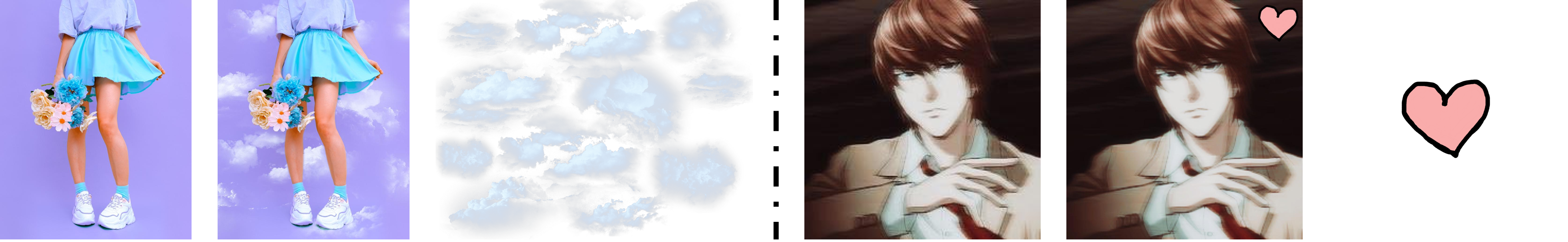}
\caption{Examples of filter-style (left) and sticker-style (right) \task~compositions. A mask is created for the filter-style composition for better visual expression.}
\vspace{-12pt}
\label{fig:styles}
\end{figure*}

\subsection{Data Source and Preprocessing}
The dataset\footnote{\textit{Picsart} is a company that complies with GDPR, CCPA, and other data protection legislation, and is collecting, storing, and processing users' data in accordance with the consent received. It also allows the users to opt-out of certain processing purposes, as requested.} is constructed from real user interactions collected over a six-month period on \textit{Picsart}, a creative image editing platform. We apply two filtering criteria to the raw data to mitigate noise and distill a expressive subset of user behaviors. First, by retaining only the top 10\% of stickers by usage frequency, we exclude niche or unpopular assets to ensure the data reflects broadly applicable content. Second, we only include compositions from shared editing workflows reused at least five times by other users. This requirement filters out edits of low quality or limited appeal, thereby retaining compositions that have demonstrated community validation and align with real user intent. Each composition action contains a host image, a sticker, and placement parameters: top-left coordinates, width, height, opacity, rotation, and an optional mask. \par

\subsection{Filter-style vs. Sticker-style}

Upon analyzing the filtered user actions, we identify two distinct composition types: \textbf{filter-style} and \textbf{sticker-style}, as illustrated in \cref{fig:styles}. Filter-style compositions involve elements used as frames, backgrounds, or visual effects, which typically span a large portion of the host image and often involve opacity or masking adjustments for better integration. In contrast, sticker-style compositions refer to standalone embellishments placed locally on specific content, usually without transparency modifications since most stickers inherently have an alpha channel. \par

\begin{figure}[htb]
    \centering
    \vspace{-8pt}
    \begin{subfigure}{0.47\textwidth}
        \centering
        \includegraphics[width=0.9\linewidth]{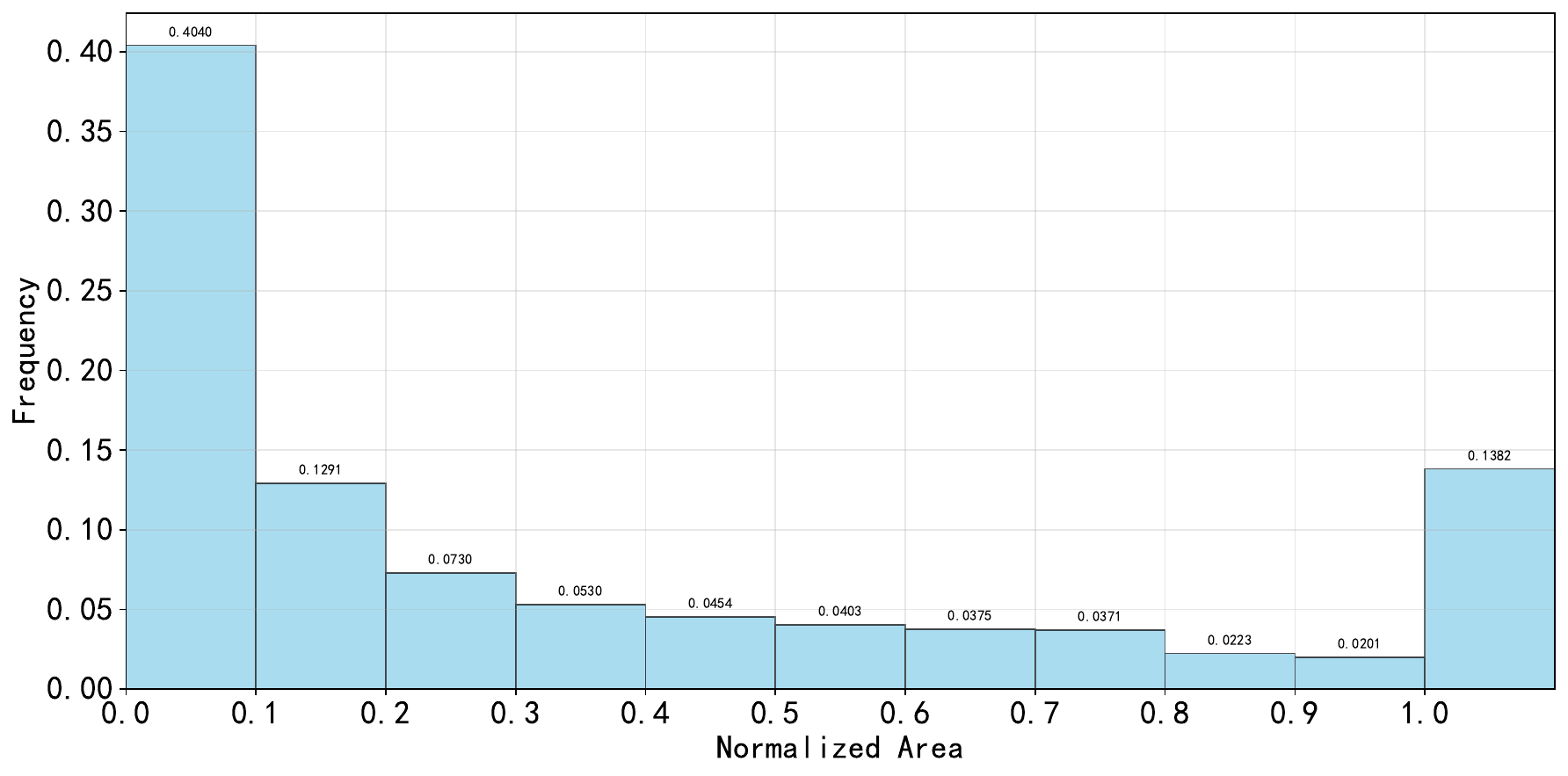}
        \caption{Distribution of the size of the added stickers}
        \label{fig:size_dist}
    \end{subfigure}
    
    \begin{subfigure}{0.47\textwidth}
        \centering
        \includegraphics[width=0.9\linewidth]{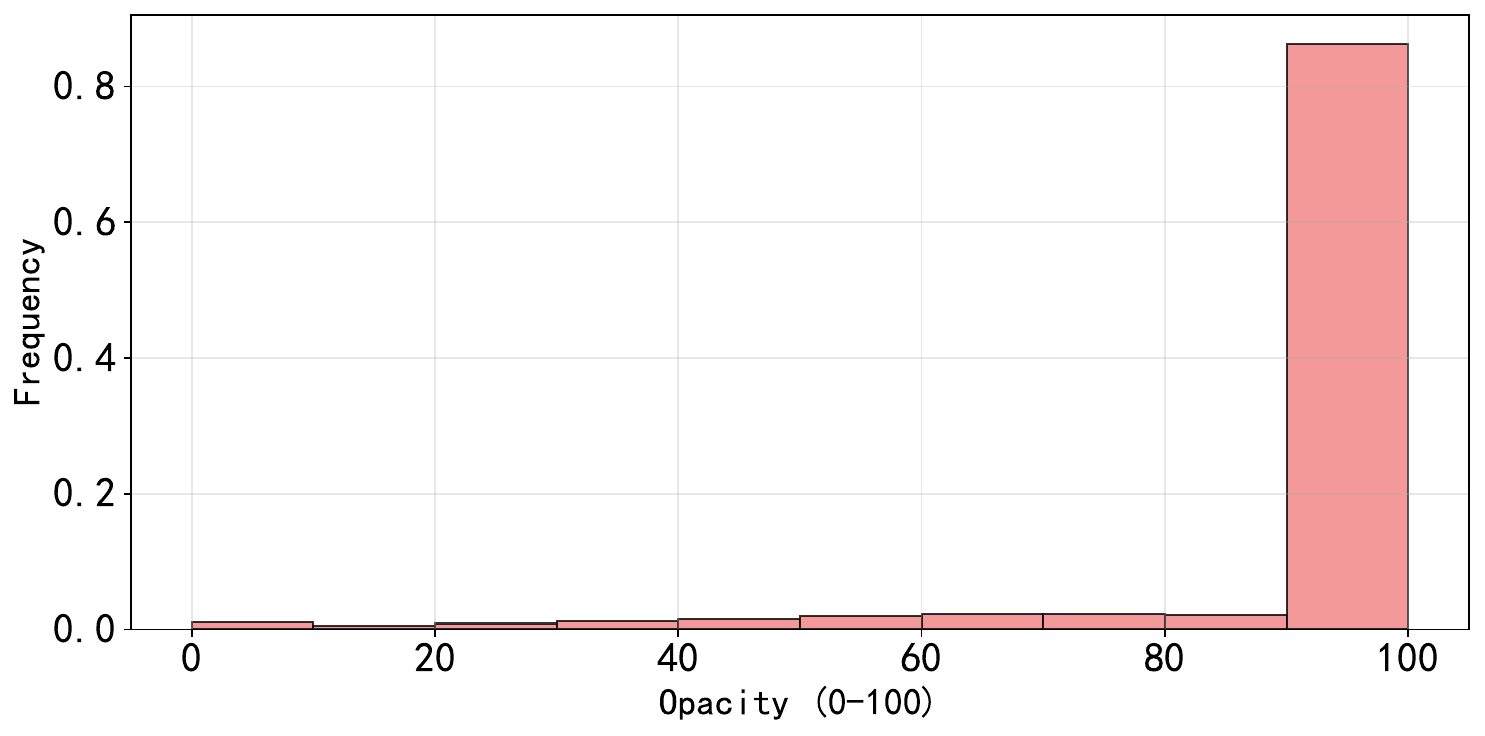}
        \caption{Distribution of the opacity of the added stickers}
        \label{fig:opa_dist}
    \end{subfigure}
    \caption{Distribution of the size and opacity of the added stickers}
    \label{fig:size_and_opa}
    \vspace{-8pt}
\end{figure}

This classification is strongly supported by the statistical properties of the dataset. The distribution of sticker size (\cref{fig:size_dist}), for instance, is notably bimodal. The primary peak, where 40.4\% of stickers occupy less than 10\% of the image area, directly corresponds to the localized nature of sticker-style additions. Conversely, the second significant peak, with 13.8\% of stickers being larger than the host image, is characteristic of the expansive overlays used in filter-style compositions. Similarly, the opacity distribution as demonstrated in \cref{fig:opa_dist} shows that while most stickers are applied with full opacity, the considerable portion with adjusted transparency often aligns with the nuanced visual integration required by the filter-style. \par

\begin{figure}[hb]
    \centering
    \begin{subfigure}{0.47\textwidth}
        \centering
        \includegraphics[width=0.9\linewidth]{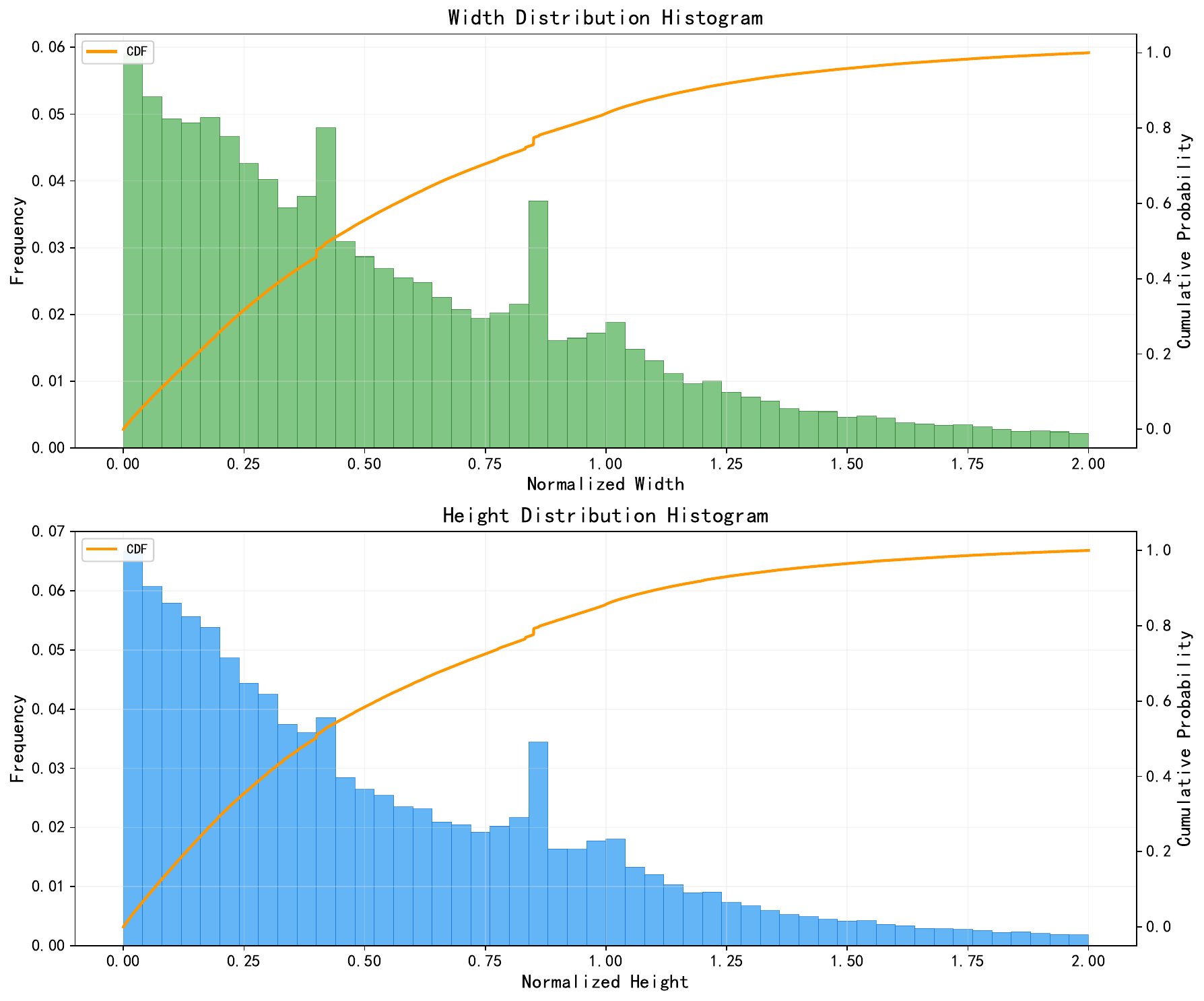}
        \caption{Histogram of the width/height of the added stickers}
        \label{fig:w_h_dist}
    \end{subfigure}
    
    \begin{subfigure}{0.47\textwidth}
        \centering
        \includegraphics[width=0.7\linewidth]{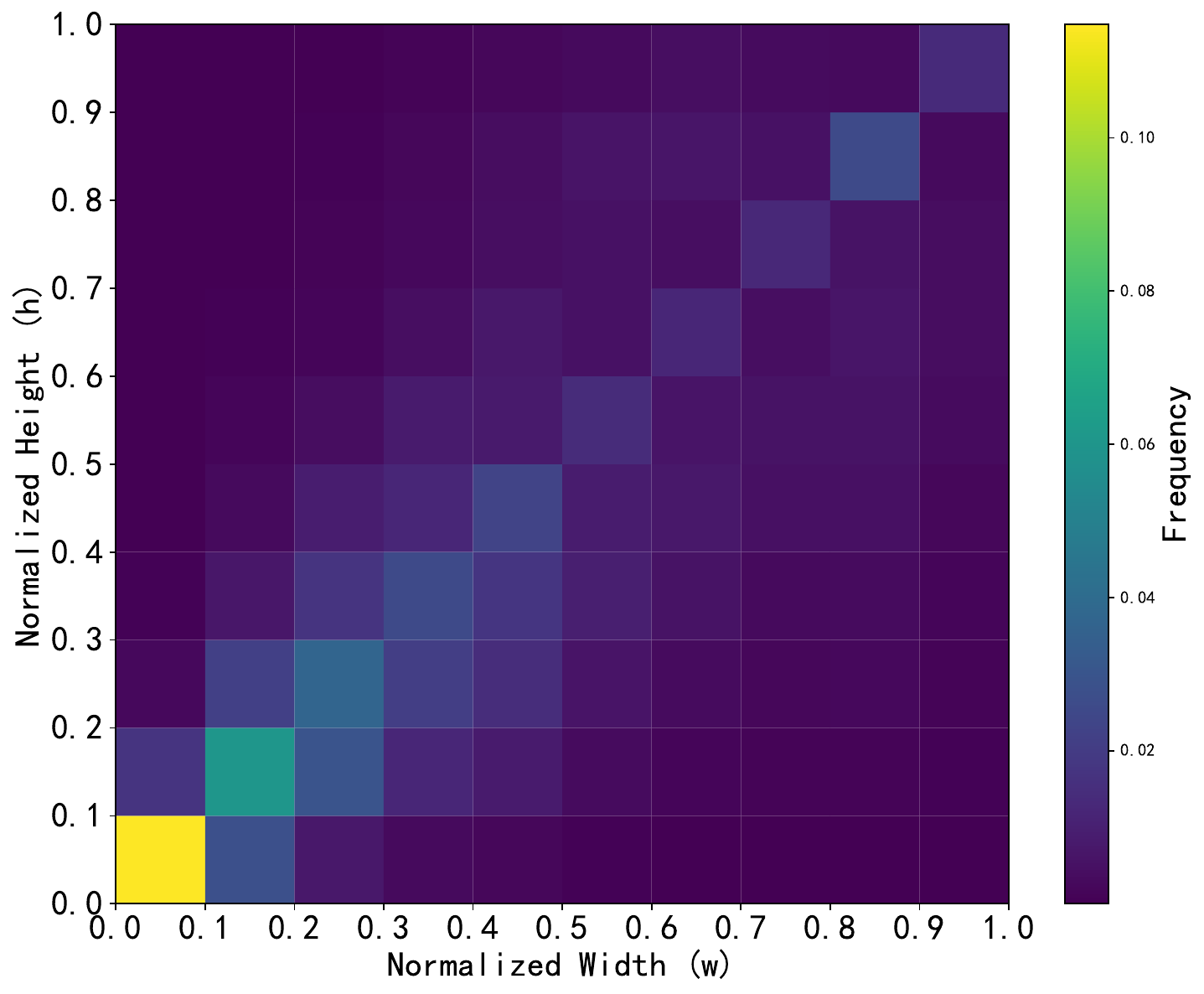}
        \caption{Heatmap of the width/height distribution}
        \label{fig:w_h_heat}
    \end{subfigure}
    \caption{Relative width and height of the added stickers}
    \label{fig:w_h}
    \vspace{-8pt}
\end{figure}

Further analysis of the final placement dimensions reveals geometric patterns that provide a strong rationale for our classification and modeling strategy. The joint distribution heatmap (\cref{fig:w_h_heat}) shows a dominant pattern: a dense concentration of data points along the diagonal ($w \approx h$) in the bottom-left corner. This subset of small and geometrically balanced additions is the statistical signature of what we define as the sticker-style. The existence of this distinct, high-density cluster justifies separating it from other, more diffuse patterns, such as the large-scale overlays that characterize the filter-style. \par 

\subsection{Dataset Labeling and Finalization}

\begin{figure}[hb]
    \centering
    \includegraphics[width=0.8\linewidth]{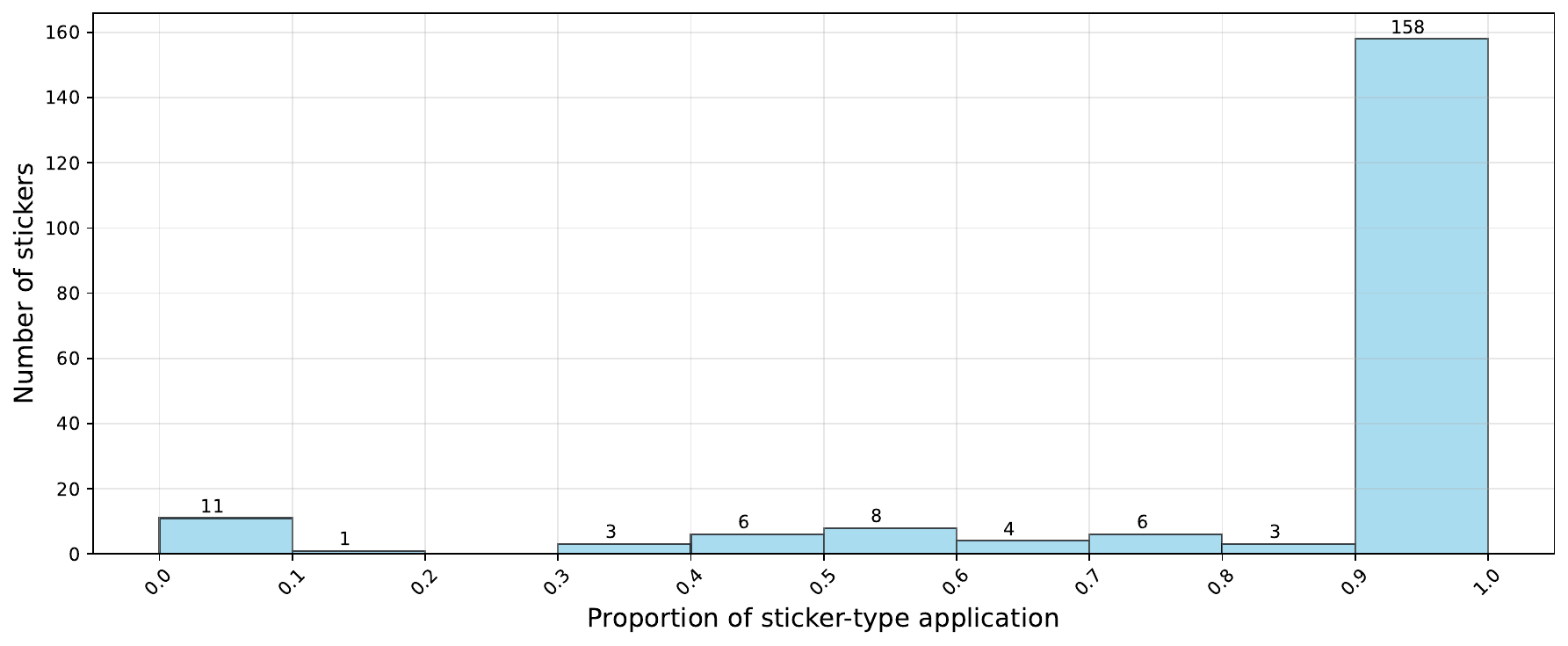}
    \caption{Distribution of Usage Consistency for Sticker Styles: the dominance of sticker-style aligns with manual annotation results.}
    \label{fig:type_dist}
    \vspace{-8pt}
\end{figure}

To better understand the usage patterns of individual stickers, we conducted another analysis of their style consistency. We randomly sampled 200 unique stickers and 50 compositions for each of them. \cref{fig:type_dist} revealed a striking finding: the distribution of the sticker-type application proportion is strongly polarized. We observed that most stickers are used almost exclusively in one of two modes: either as sticker-style (large proportion) or filter-style (small proportion), with very few exhibiting ambiguous usage. \par

This empirical evidence leads us to conclude that the composition type is a stable, intrinsic attribute of the sticker itself, rather than a context-dependent choice. Leveraging this data-driven conclusion, we proceed to label our dataset at the sticker level. We first identify candidate filter-style stickers by applying a heuristic: stickers that cover more than 50\% of the host image in at least 50\% of their uses. These candidates are then manually annotated to finalize the labels, resulting in 3,672 filter-style stickers. All remaining stickers are subsequently classified as sticker-style, yielding 69,269 instances.

The data-driven distinctions also inform the design of our prediction targets. For the sticker-style, where user intent is clearly expressed through choices of position and size, the essential task is to predict these very parameters. Conversely, for filter-style compositions, where the geometry is often trivial (e.g., covering the entire image), the user's creative decision shifts from `placement' to `blending'. Therefore, predicting global attributes like opacity or masking becomes is more important for this category. \par
To balance the dataset, we randomly sample up to 300 usage instances per sticker, creating a final set of 1,881,169 samples. The dataset is split at the sticker level into training (90\%), validation (5\%), and test (5\%) sets. This ensures that no sticker appears across splits, enabling a robust evaluation of model generalization to unseen stickers. \par

\section{Method}
\label{sec:method}

\subsection{Overall Design}
\begin{figure}[htb]
\centering
\includegraphics[width=0.45\textwidth]{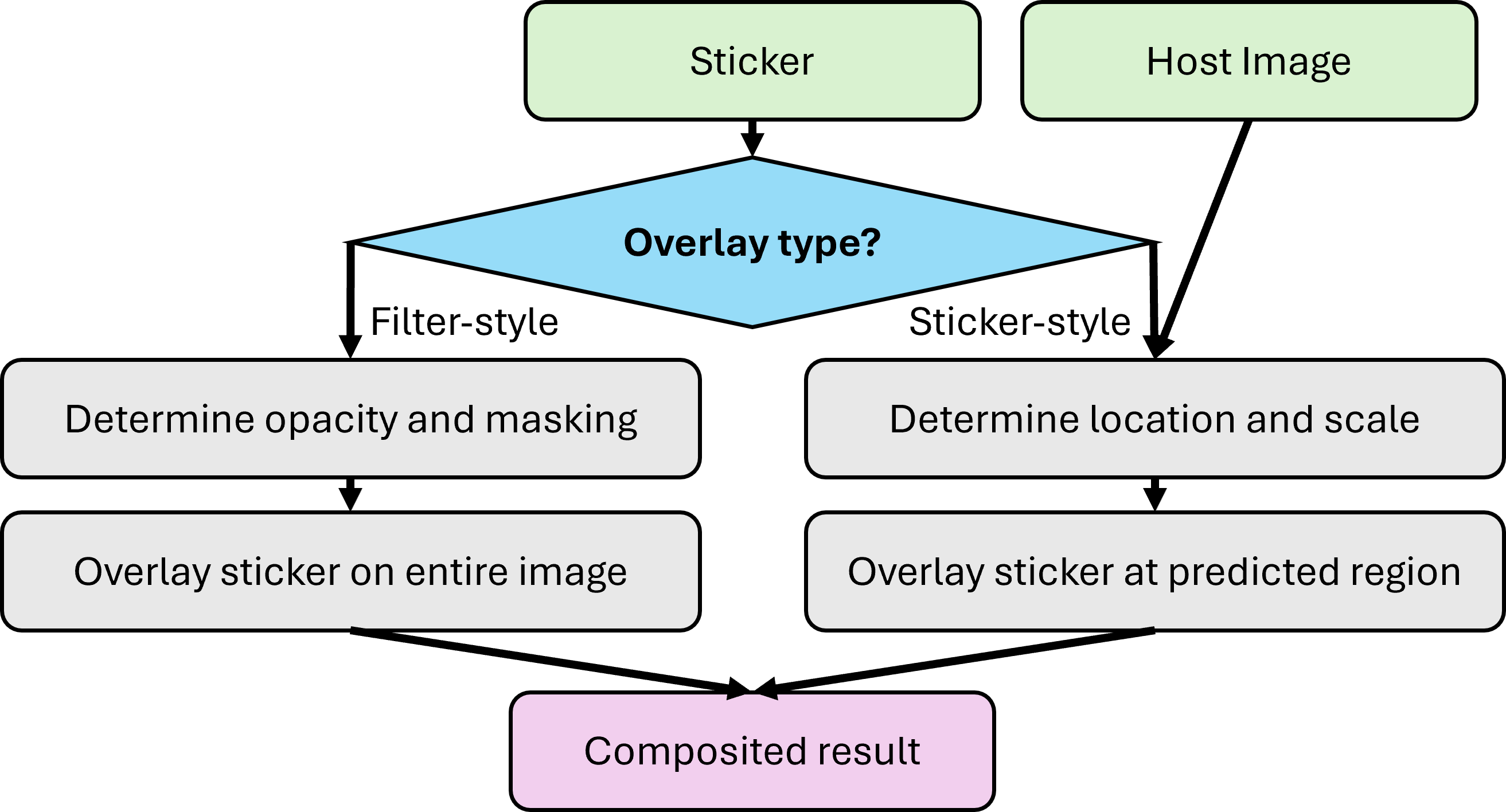}
\caption{\netname~pipeline consisting of a type classifier and a placement predictor}
\label{fig:overall}
\vspace{-8pt}
\end{figure}

Based on the analysis in \cref{sec:dataset}, we design a two-stage model, \textbf{\netname}, tailored to the characteristics of the task. The first stage predicts the composition type, followed by type-specific prediction of placement parameters. As in \cref{fig:overall}, the pipeline starts with a classifier that identifies whether the composition is filter-style or sticker-style. For filter-style, the model applies a global overlay across the host image and predicts whether opacity adjustment or masking is required. For sticker-style, a placement predictor conditioned on both the sticker and the host image estimates the location and scale of the placement. Details of the type classifier and placement predictor are described in the following sections.

\subsection{Type Classifier}

\begin{figure}[htb]
\centering
\vspace{-8pt}
\includegraphics[width=0.47\textwidth]{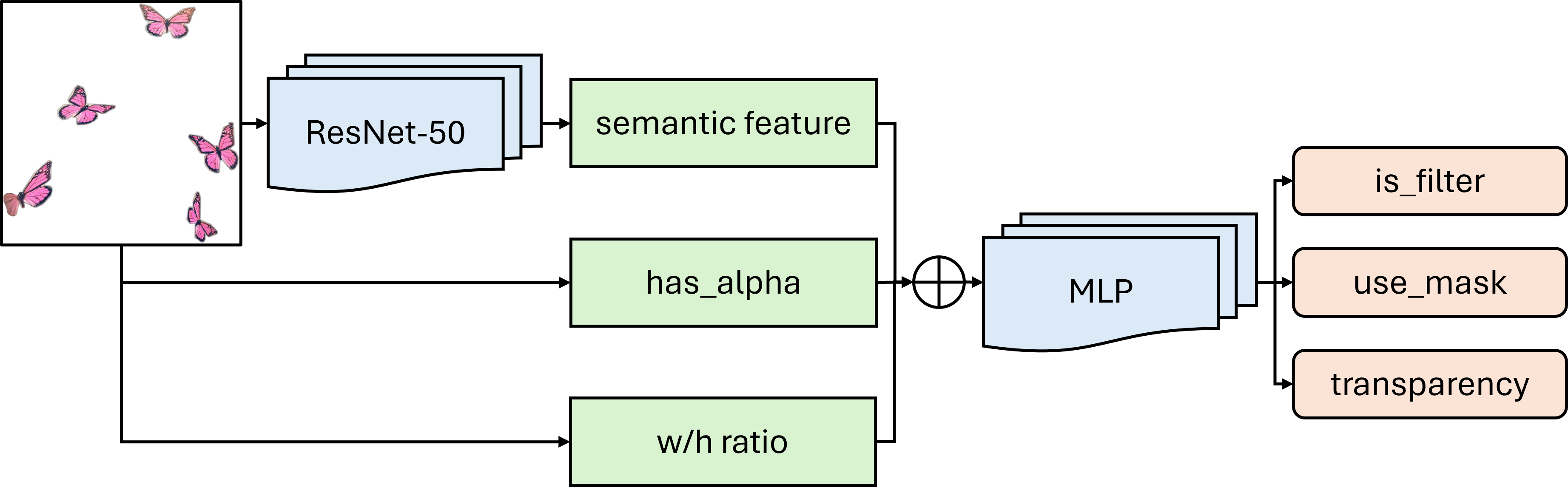}
\caption{Architecture of the type classifier}
\label{fig:classifier}
\vspace{-8pt}
\end{figure}

As shown in \cref{fig:classifier}, the type classifier consists of a ResNet-50 \cite{he2016deep} backbone adapted for RGBA inputs by modifying the first convolutional layer to accept four channels. A binary flag \textit{has\_alpha} is set to 1 if the sticker includes an alpha channel, and 0 otherwise; in the latter case, a constant alpha channel of 255 is appended. The normalized aspect ratio, defined in \cref{eq:whratio}, is zero for square stickers and increases as the shape becomes more elongated. The semantic feature extracted by ResNet-50 is concatenated with \textit{has\_alpha} and the aspect ratio to form a joint feature vector, which is passed through a three-layer MLP to predict three outputs: \textit{is\_filter}, \textit{use\_mask}, and \textit{transparency}.

\begin{equation}
\label{eq:whratio}
    r = \left(1 - \frac{\min(w, h)}{\max(w, h)}\right)^2
\end{equation}

\subsection{Placement Predictor}
\begin{figure*}[htb]
\centering
\includegraphics[width=0.9\textwidth]{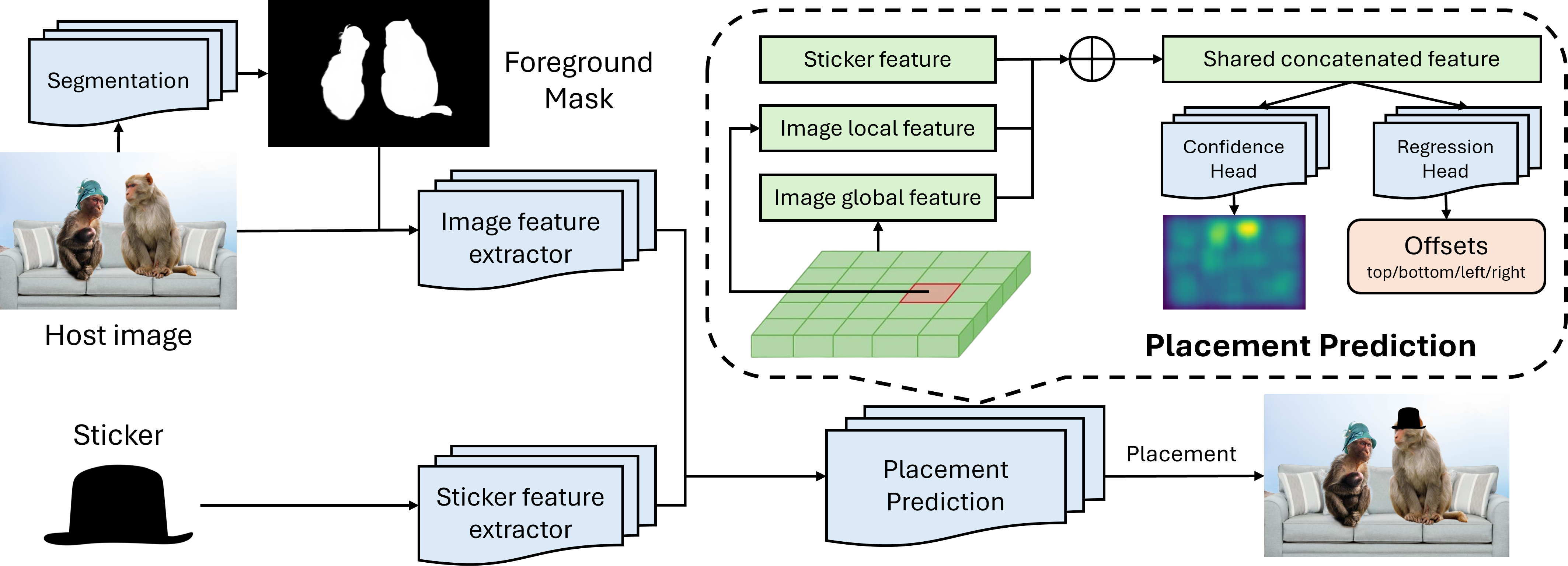}
\caption{Architecture of the Placement Predictor, including multi-scale feature extraction, region-level confidence estimation, and offset regression for sticker positioning}
\label{fig:general}
\vspace{-12pt}
\end{figure*}

If a sticker is classified as sticker-style, a placement predictor is invoked to estimate its optimal location and scale on the host image. As in \cref{fig:general}, the model consists of two components: feature extraction and placement prediction.

Sticker and host features are extracted using separate ResNet-50 backbones. The sticker branch processes a 4-channel RGBA input, while the host branch uses a 5-channel input comprising its RGBA channels and a foreground mask. The specific choice of segmentation model does not appear to be a critical factor. Our supporting experiments show, for example, that swapping our default model with Mask R-CNN \cite{he2017mask} leads to a negligible performance change (see \cref{tab:ablation}). For the ease of reproducibility, the mask used throughout our experiments is generated by a pretrained DeepLabV3 \cite{chen2017rethinking} model. Features from layers 2 and 3 of the backbones are used for multi-scale dense prediction. At each spatial location in the host feature map, a local feature is concatenated with global context vectors from both the sticker and the host image. Following the common design pattern, this combined representation is passed to two convolutional heads: one predicts placement confidence, and the other regresses relative scale along x and y. The final placement corresponds to the location with the highest confidence.

\subsection{Implementation Details}
During training, the input sticker for the type classifier is resized to 300x300 and randomly cropped to 256x256. A 2048-dimensional feature is extracted using ResNet-50 and concatenated with the binary \textit{has\_alpha} indicator and the normalized aspect ratio, resulting in a 2050-dimensional vector. This vector is passed through a 3-layer MLP with sigmoid activation to produce three probabilities: whether the sticker is filter-style, whether a foreground mask should be applied, and whether the opacity should be reduced. At inference time, thresholds are applied to these probabilities to obtain binary decisions. Stickers classified as sticker-style are passed to the placement predictor. Filter-style stickers are directly composited by resizing them to match the host image, with optional masking and opacity adjustment applied according to the predictions. \par

For the placement predictor, both the host image and sticker are resized to 256x256. We use feature maps from layers 2 and 3 of the ResNet-50 backbones, with spatial resolutions of 32x32 and 16x16, respectively, yielding 1280 candidate anchors. The loss function consists of two components: a binary classification loss that determines whether each anchor corresponds to a plausible placement, and a regression loss that refines the offset and scale of the placement. Classification loss is computed using binary cross-entropy over all anchors, while regression loss is applied only to positive anchors that fall within the ground-truth sticker region in the original image. \par

We adopt Distance-IoU (DIoU)~\cite{zheng2020distance} loss as the regression loss, which extends standard IoU by penalizing the distance between box centers and incorporating scale alignment as defined in \cref{eq:diou}. $B$ and $B^{gt}$ represent the predicted and ground-truth boxes, $b$ and $b^{gt}$ are their center points, $\rho$ is the Euclidean distance between centers, and $c$ is the diagonal length of the smallest enclosing box. Compared with standard IoU, DIoU provides informative gradients even when there is little or no overlap, enabling more stable optimization.
\vspace{-7pt}
\begin{equation}
\label{eq:diou}
\text{DIoU}(B, B^{gt}) = \text{IoU}(B, B^{gt}) - \frac{\rho^2(b, b^{gt})}{c^2}
\end{equation}

The overall loss function is given in \cref{eq:loss}. Here, $y_i \in \{0, 1\}$ and $p_i$ denote the ground-truth label and the predicted confidence score of the $i$-th anchor; $B_i$, $B^{gt}_i$ are the predicted and ground-truth boxes. The regression loss is averaged over positive anchors, and the balancing weight $\lambda$ is set to 3 in our experiments.

\vspace{-7pt}
\begin{equation}
\label{eq:loss}
\mathcal{L} = \frac{\sum_{i=1}^{N} \mathcal{L}_{\text{BCE}}(p_i, y_i)}{N} 
+ \lambda \cdot \frac{\sum_{i=1}^{N} y_i \cdot (1 - \text{DIoU}(B_i, B^{gt}_i))}{\sum_{i=1}^{N} y_i}
\end{equation}
\vspace{-8pt}

The model is lightweight and efficient. On a single NVIDIA GeForce RTX 2060 GPU without batch processing, the type classifier and placement predictor take 45ms/92ms on average, respectively. The codebase for \netname~model training and evaluation along with model weights is provided as supplementary material. 
\section{Evaluation}
\label{sec:exp}

As our task prioritizes subjective goals like aesthetic quality and expressiveness, our evaluation is centered on a user study to directly measure human perception. The results, based on 500 samples from the test set, are complemented by a quantitative analysis that serves as a proxy for placement quality and is used for ablation studies.

\subsection{User Study}  
To directly assess alignment with user preference, our primary metric, we conducted a user study comparing \netname~against several baselines:

\begin{itemize}
    \item \textbf{User History}: The sticker is placed according to the original user action.
    \item \textbf{Center}: The sticker is placed at the center of the host image, preserving its aspect ratio and occupying 1/9 of the image area. \textbf{It is the default behavior in most editing platforms.}
    \item \textbf{Random}: The sticker is placed at a random location, with random aspect ratio scaling (50\%-200\% that of the original sticker) and area (1/25-1x that of the host image).
    \item \textbf{GracoNet}: Placement predicted by pretrained GracoNet~\cite{zhou2022learning}\footnote{As the annotations required to train GracoNet cannot be derived from our dataset, we use the publicly released pretrained model for evaluation.}, one of the best open-sourced conventional image composition model.
    \item \textbf{\netname}: Placement predicted by our model.
\end{itemize}

\begin{figure}[htb]
\centering
\includegraphics[width=0.45\textwidth]{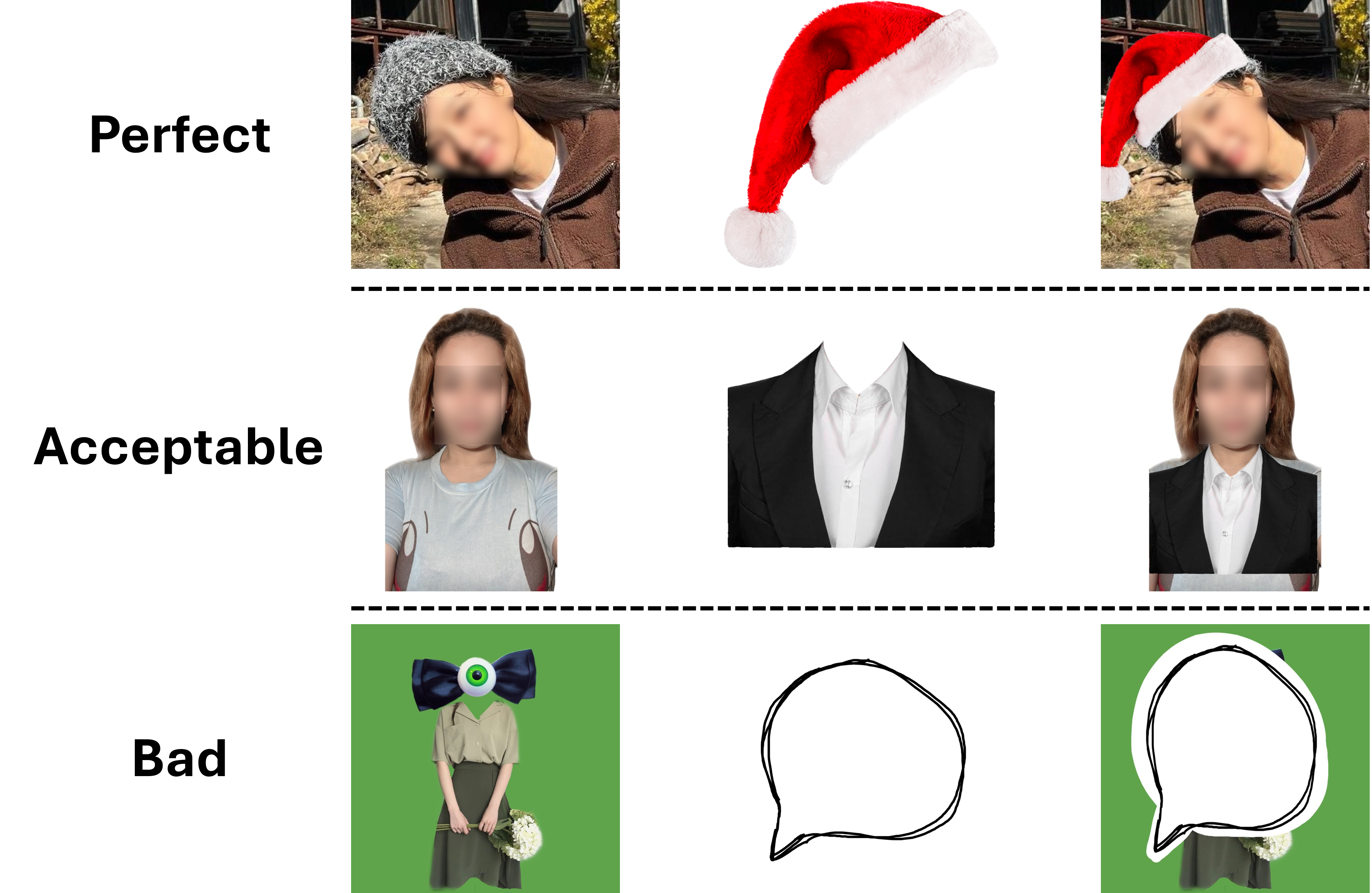}
\caption{Examples from the user study. Each row corresponds to a different rating category. From left to right: input image, sticker, and composite. Faces are blurred for privacy.}
\label{fig:rating}
\vspace{-6pt}
\end{figure}

Each output was evaluated by five independent annotators and assigned to one of four quality classes (illustrated in \cref{fig:rating}) based on majority voting: \textbf{Perfect}, \textbf{Acceptable}, \textbf{Bad}, or \textbf{Invalid}.

\begin{itemize}
    \item \textbf{Perfect:} The placement closely aligns with the expectation of the annotators, requiring minimal or no further edits.
    \item \textbf{Acceptable:} Reasonable placement, minor adjustments in location/scale may be needed.
    \item \textbf{Bad:} Clearly misaligned with the expectations of the annotators.
    \item \textbf{Invalid:} No majority agreement, sticker not visible, or content deemed NSFW.
\end{itemize}

\begin{figure}[htb]
\centering
\vspace{-12pt}
\includegraphics[width=0.47\textwidth]{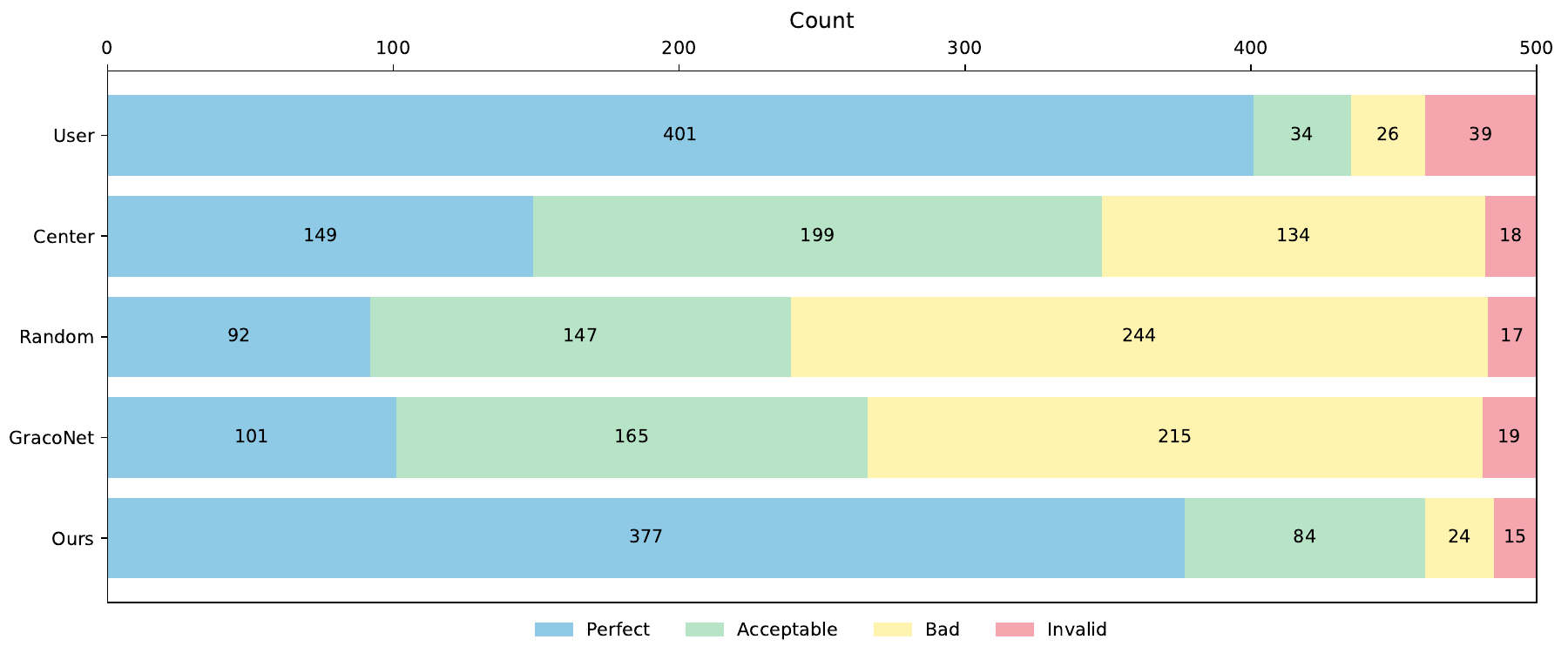}
\caption{Distribution of quality ratings for different methods.}
\vspace{-8pt}
\label{fig:userstudy}
\end{figure}

\cref{fig:userstudy} summarizes the results. User history placements were rated as \textit{Perfect} in 86.8\% of cases, demonstrating the general reliability of human edits. However, they also accounted for the largest proportion of samples without annotator consensus, and 5\% were rated as \textit{Bad}. The default center placement was surprisingly effective, receiving at least an \textit{Acceptable} rating in nearly 70\% of cases. Further analysis reveals that in approximately half of the samples, the sticker was attached to the most salient object in the host image, which was typically centered. This partially explains the effectiveness of center-based strategies. \par
Random placement performed poorly, with fewer than half of the results rated \textit{Acceptable} or better. Although GracoNet slightly outperformed random placement, its overall performance remained poor in this setting due to two key factors: (1) the semantics of many stickers fall outside the distribution of natural image datasets, and (2) unlike traditional foreground object placement, stickers can function as decorative or auxiliary elements. \par
\netname~achieved strong performance: while it received slightly fewer \textit{Perfect} ratings than user history, it obtained at least an \textit{Acceptable} rating on more samples. Most importantly, it significantly outperformed other baselines. To better understand remaining limitations, we manually analyzed the 24 samples where \netname~received a \textit{Bad} rating. Among them, 5 cases required advanced processing for stickers like cropping, which are not supported by the current framework. In 6 cases, the sticker was semantically or stylistically unsuitable, as evidenced by the \textit{Bad} ratings of the corresponding user placements. Another 6 failures were caused by incorrect type classification, and the remaining 7 cases were due to inaccurate placement predictions despite correct type classification. These findings suggest that while \netname~is generally reliable, further improvements are possible through better editing support, type classification, and placement robustness.

\subsection{Quantitative Analysis}  
To complement the user study and provide a quantitative measure for ablation, we adopt DIoU as a proxy for geometric placement quality. \cref{tab:ablation} presents the ablation study and a comparison with the baselines introduced above. The ablation results demonstrate that both the type classifier (TC) and the use of DIoU loss provide significant gains. Furthermore, \netname~outperforms all baselines by a wide margin on this metric. The overall scores align well with the human ratings from the user study, supporting DIoU as a reasonable proxy for this task.

\begin{table}[htb]
\begin{center}
\vspace{-6pt}
\begin{tabular}{@{}lc@{}}
\toprule
Method & DIoU $\uparrow$ \\
\midrule
Center & -0.0335 \\
Random & -0.1471 \\
GracoNet & -0.0983 \\
\midrule
\textbf{\netname} & \textbf{0.2050} \\
\midrule
\netname~using Mask R-CNN & 0.2041 \\
\netname~using native IoU loss & 0.1812 \\
\netname~w/o type classifier (TC) & 0.1362 \\
\netname~w/o TC and using native IoU loss & 0.1158 \\
\bottomrule
\end{tabular}
\end{center}
\vspace{-12pt}
\caption{Ablation and baseline comparison. Higher scores indicate better spatial alignment with ground-truth placements.}
\label{tab:ablation}
\end{table}

\vspace{-16pt}
\section{Conclusion}

In this work, we introduce \task, a new formulation of image composition centered on expressiveness and user intent. To address it, we propose \netname, a lightweight two-stage framework that classifies sticker type and predicts type-specific placement, making it suitable for practical on-device applications. Trained on real-world popular editing data, \netname~outperforms standard baselines and closely reflects human behavior, highlighting the value of learning from massive authentic creative actions.

Despite its strong performance, \netname~has several limitations. The current framework does not support advanced operations such as rotation, cropping, or layering, which are often present in real editing workflows. In addition, the model relies on user-generated placements, which may contain noise or suboptimal decisions. The training set is also restricted to frequently used stickers, which may limit generalization to rarer styles. \par

Future work will focus on three key directions. First, we aim to extend the framework to advanced operations. Second, we will explore techniques like human-in-the-loop refinement to improve data quality and generalization. Finally, a long-term direction is developing personalized models that adapt to the unique aesthetic preferences of individual users.

{
    \small
    \bibliographystyle{ieeenat_fullname}
    \bibliography{main}

@String(ICME = {Int. Conf. Multimedia and Expo})

@String(AAAI = {AAAI})

@String(ICME  =	{ICME})

@article{lee2018context,
  title={Context-aware synthesis and placement of object instances},
  author={Lee, Donghoon and Liu, Sifei and Gu, Jinwei and Liu, Ming-Yu and Yang, Ming-Hsuan and Kautz, Jan},
  journal={Advances in neural information processing systems},
  volume={31},
  year={2018}
}

@inproceedings{wang2023gan,
  title={Ca-gan: Object placement via coalescing attention based generative adversarial network},
  author={Wang, Yibin and Feng, Yuchao and Wu, Jie and Xu, Honghui and Zheng, Jianwei},
  booktitle={2023 IEEE International Conference on Multimedia and Expo (ICME)},
  pages={2375--2380},
  year={2023},
  organization={IEEE}
}

@inproceedings{zhou2022learning,
  title={Learning object placement via dual-path graph completion},
  author={Zhou, Siyuan and Liu, Liu and Niu, Li and Zhang, Liqing},
  booktitle={European Conference on Computer Vision},
  pages={373--389},
  year={2022},
  organization={Springer}
}

@article{zhou2022sac,
  title={SAC-GAN: Structure-aware image composition},
  author={Zhou, Hang and Ma, Rui and Zhang, Ling-Xiao and Gao, Lin and Mahdavi-Amiri, Ali and Zhang, Hao},
  journal={IEEE Transactions on Visualization and Computer Graphics},
  year={2022},
  publisher={IEEE}
}

@article{turgutlu2022layoutbert,
  title={Layoutbert: Masked language layout model for object insertion},
  author={Turgutlu, Kerem and Sharma, Sanat and Kumar, Jayant},
  journal={arXiv preprint arXiv:2205.00347},
  year={2022}
}

@article{ye2023efficient,
  title={Efficient object placement via ftopnet},
  author={Ye, Guosheng and Wang, Jianming and Yang, Zizhong},
  journal={Electronics},
  volume={12},
  number={19},
  pages={4106},
  year={2023},
  publisher={MDPI}
}

@inproceedings{zhu2023topnet,
  title={Topnet: Transformer-based object placement network for image compositing},
  author={Zhu, Sijie and Lin, Zhe and Cohen, Scott and Kuen, Jason and Zhang, Zhifei and Chen, Chen},
  booktitle={Proceedings of the IEEE/CVF Conference on Computer Vision and Pattern Recognition},
  pages={1838--1847},
  year={2023}
}

@article{niu2022fast,
  title={Fast object placement assessment},
  author={Niu, Li and Liu, Qingyang and Liu, Zhenchen and Li, Jiangtong},
  journal={arXiv preprint arXiv:2205.14280},
  year={2022}
}

@inproceedings{liu2024conditional,
  title={Conditional Transformation Diffusion for Object Placement},
  author={Liu, Jiacheng and Wei, Shida and Ma, Rui},
  booktitle={Proceedings of the 2024 7th International Conference on Image and Graphics Processing},
  pages={363--368},
  year={2024}
}

@inproceedings{zhang2020learning,
  title={Learning object placement by inpainting for compositional data augmentation},
  author={Zhang, Lingzhi and Wen, Tarmily and Min, Jie and Wang, Jiancong and Han, David and Shi, Jianbo},
  booktitle={Computer Vision--ECCV 2020: 16th European Conference, Glasgow, UK, August 23--28, 2020, Proceedings, Part XIII 16},
  pages={566--581},
  year={2020},
  organization={Springer}
}

@inproceedings{zhu2022gala,
  title={Gala: Toward geometry-and-lighting-aware object search for compositing},
  author={Zhu, Sijie and Lin, Zhe and Cohen, Scott and Kuen, Jason and Zhang, Zhifei and Chen, Chen},
  booktitle={European Conference on Computer Vision},
  pages={676--692},
  year={2022},
  organization={Springer}
}

@inproceedings{lin2018st,
  title={St-gan: Spatial transformer generative adversarial networks for image compositing},
  author={Lin, Chen-Hsuan and Yumer, Ersin and Wang, Oliver and Shechtman, Eli and Lucey, Simon},
  booktitle={Proceedings of the IEEE conference on computer vision and pattern recognition},
  pages={9455--9464},
  year={2018}
}

@article{liu2021opa,
  title={Opa: object placement assessment dataset},
  author={Liu, Liu and Liu, Zhenchen and Zhang, Bo and Li, Jiangtong and Niu, Li and Liu, Qingyang and Zhang, Liqing},
  journal={arXiv preprint arXiv:2107.01889},
  year={2021}
}

@inproceedings{song2023objectstitch,
  title={Objectstitch: Object compositing with diffusion model},
  author={Song, Yizhi and Zhang, Zhifei and Lin, Zhe and Cohen, Scott and Price, Brian and Zhang, Jianming and Kim, Soo Ye and Aliaga, Daniel},
  booktitle={Proceedings of the IEEE/CVF Conference on Computer Vision and Pattern Recognition},
  pages={18310--18319},
  year={2023}
}

@inproceedings{ma2024directed,
  title={Directed diffusion: Direct control of object placement through attention guidance},
  author={Ma, Wan-Duo Kurt and Lahiri, Avisek and Lewis, John P and Leung, Thomas and Kleijn, W Bastiaan},
  booktitle={Proceedings of the AAAI Conference on Artificial Intelligence},
  volume={38},
  number={5},
  pages={4098--4106},
  year={2024}
}

@inproceedings{zhang2018unreasonable,
  title={The unreasonable effectiveness of deep features as a perceptual metric},
  author={Zhang, Richard and Isola, Phillip and Efros, Alexei A and Shechtman, Eli and Wang, Oliver},
  booktitle={Proceedings of the IEEE conference on computer vision and pattern recognition},
  pages={586--595},
  year={2018}
}

@inproceedings{he2016deep,
  title={Deep residual learning for image recognition},
  author={He, Kaiming and Zhang, Xiangyu and Ren, Shaoqing and Sun, Jian},
  booktitle={Proceedings of the IEEE conference on computer vision and pattern recognition},
  pages={770--778},
  year={2016}
}

@article{chen2017rethinking,
  title={Rethinking atrous convolution for semantic image segmentation},
  author={Chen, Liang-Chieh and Papandreou, George and Schroff, Florian and Adam, Hartwig},
  journal={arXiv preprint arXiv:1706.05587},
  year={2017}
}

@inproceedings{zheng2020distance,
  title={Distance-IoU loss: Faster and better learning for bounding box regression},
  author={Zheng, Zhaohui and Wang, Ping and Liu, Wei and Li, Jinze and Ye, Rongguang and Ren, Dongwei},
  booktitle={Proceedings of the AAAI conference on artificial intelligence},
  volume={34},
  number={07},
  pages={12993--13000},
  year={2020}
}

@inproceedings{kim2025orida,
  title={ORIDa: Object-centric Real-world Image Composition Dataset},
  author={Kim, Jinwoo and Han, Sangmin and Jeong, Jinho and Choi, Jiwoo and Kim, Dongyeoung and Kim, Seon Joo},
  booktitle={Proceedings of the Computer Vision and Pattern Recognition Conference},
  pages={3051--3060},
  year={2025}
}

@inproceedings{qin2024think,
  title={Think before placement: Common sense enhanced transformer for object placement},
  author={Qin, Yaxuan and Xu, Jiayu and Wang, Ruiping and Chen, Xilin},
  booktitle={European Conference on Computer Vision},
  pages={35--50},
  year={2024},
  organization={Springer}
}

@article{zhang2023magicbrush,
  title={Magicbrush: A manually annotated dataset for instruction-guided image editing},
  author={Zhang, Kai and Mo, Lingbo and Chen, Wenhu and Sun, Huan and Su, Yu},
  journal={Advances in Neural Information Processing Systems},
  volume={36},
  pages={31428--31449},
  year={2023}
}

@article{hui2024hq,
  title={Hq-edit: A high-quality dataset for instruction-based image editing},
  author={Hui, Mude and Yang, Siwei and Zhao, Bingchen and Shi, Yichun and Wang, Heng and Wang, Peng and Zhou, Yuyin and Xie, Cihang},
  journal={arXiv preprint arXiv:2404.09990},
  year={2024}
}

@inproceedings{sushko2025realedit,
  title={Realedit: Reddit edits as a large-scale empirical dataset for image transformations},
  author={Sushko, Peter and Bharadwaj, Ayana and Lim, Zhi Yang and Ilin, Vasily and Caffee, Ben and Chen, Dongping and Salehi, Mohammadreza and Hsieh, Cheng-Yu and Krishna, Ranjay},
  booktitle={Proceedings of the Computer Vision and Pattern Recognition Conference},
  pages={13403--13413},
  year={2025}
}

@inproceedings{kawar2023imagic,
  title={Imagic: Text-based real image editing with diffusion models},
  author={Kawar, Bahjat and Zada, Shiran and Lang, Oran and Tov, Omer and Chang, Huiwen and Dekel, Tali and Mosseri, Inbar and Irani, Michal},
  booktitle={Proceedings of the IEEE/CVF conference on computer vision and pattern recognition},
  pages={6007--6017},
  year={2023}
}

@article{yang2024editworld,
  title={Editworld: Simulating world dynamics for instruction-following image editing},
  author={Yang, Ling and Zeng, Bohan and Liu, Jiaming and Li, Hong and Xu, Minghao and Zhang, Wentao and Yan, Shuicheng},
  journal={arXiv preprint arXiv:2405.14785},
  year={2024}
}

@article{ge2024seed,
  title={Seed-data-edit technical report: A hybrid dataset for instructional image editing},
  author={Ge, Yuying and Zhao, Sijie and Li, Chen and Ge, Yixiao and Shan, Ying},
  journal={arXiv preprint arXiv:2405.04007},
  year={2024}
}

@inproceedings{he2017mask,
  title={Mask r-cnn},
  author={He, Kaiming and Gkioxari, Georgia and Doll{\'a}r, Piotr and Girshick, Ross},
  booktitle={Proceedings of the IEEE international conference on computer vision},
  pages={2961--2969},
  year={2017}
}

@article{lu2024can,
  title={Can Vision-Language Models Replace Human Annotators: A Case Study with CelebA Dataset},
  author={Lu, Haoming and Zhong, Feifei},
  journal={arXiv preprint arXiv:2410.09416},
  year={2024}
}

@article{lu2025trueskin,
  title={TrueSkin: Towards Fair and Accurate Skin Tone Recognition and Generation},
  author={Lu, Haoming},
  journal={arXiv preprint arXiv:2509.10980},
  year={2025}
}
}

\end{document}